\documentclass{article}

\usepackage{microtype}
\usepackage{graphicx}
\usepackage{amsmath}
\usepackage{subfigure}
\usepackage{booktabs} 
\usepackage{amsmath}
\usepackage{tikz}
\usepackage{xltabular}
\usepackage{natbib}
\usepackage{cite} 
\usepackage{stackrel}
\usepackage{pifont}
\usepackage{xcolor}
\newcommand{\cmark}{\textcolor{black}{\ding{51}}}
\newcommand{\xmark}{\textcolor{black}{\ding{55}}}
\usepackage{amsfonts}

\usepackage{multirow}

\usepackage{hyperref}

\usepackage[accepted]{icml2023}
\icmltitlerunning{Half-Hop: A graph upsampling approach for slowing down message passing}

\begin{document}

\twocolumn[
\icmltitle{Half-Hop: A graph upsampling approach for slowing down message passing}




\begin{icmlauthorlist}
\icmlauthor{Mehdi Azabou}{ga}
\icmlauthor{Venkataramana Ganesh}{ga}
\icmlauthor{Shantanu Thakoor}{dm}
\icmlauthor{Chi-Heng Lin}{ga}
\icmlauthor{Lakshmi Sathidevi}{ga}\\
\icmlauthor{Ran Liu}{ga}
\icmlauthor{Michal Valko}{dm}
\icmlauthor{Petar Veličković}{dm}
\icmlauthor{Eva L. Dyer}{ga}
\end{icmlauthorlist}

\icmlaffiliation{ga}{Georgia Tech}
\icmlaffiliation{dm}{DeepMind}

\icmlcorrespondingauthor{Mehdi Azabou and Eva Dyer}{\{mazabou, evadyer\}@gatech.edu}

\icmlkeywords{Machine Learning, ICML}

\vskip 0.3in
]



\printAffiliationsAndNotice{} 

\begin{abstract}
Message passing neural networks have shown a lot of success on graph-structured data. However, there are many instances where message passing can lead to over-smoothing or fail when neighboring nodes belong to different classes.
In this work, we introduce a simple yet general framework for improving learning in message passing neural networks.  
Our approach essentially upsamples edges in the original graph by adding ``slow nodes'' at each edge that can mediate communication between a source and a target node. Our method only modifies the input graph, making it plug-and-play and easy to use with existing models. To understand the benefits of slowing down message passing, we provide theoretical and empirical analyses. We report results on several supervised and self-supervised benchmarks, and show improvements across the board, notably in heterophilic conditions where adjacent nodes are more likely to have different labels. Finally, we show how our approach can be used to generate augmentations for self-supervised learning, where slow nodes are randomly introduced into different edges in the graph to generate multi-scale views with variable path lengths.
\end{abstract}

\section{Introduction}
Graph neural networks (GNN) are now a widely used class of artificial neural networks, with applications in recommender systems \cite{ying2018graph}, drug discovery \cite{stokes2020deep,gaudelet2021utilizing}, and much more \cite{monti2017geometric,cui2019traffic,ktena2017distance}. However, because graphs are highly variable, building general approaches for learning on graphs that work robustly on many different problems has been a major challenge.  

Most GNNs rely on message passing (MP) that leverages the graph structure to perform inference \cite{gilmer2017neural}. Message passing, while intuitive and simple, can also be limiting in some cases \cite{alon2021on,ricciflow,oono2020graph}. This is especially true when working with complex and heterophilic  graphs where nodes from different classes are connected \cite{heterophily}, and in cases where the degree distributions and connectivity is varied throughout the network \cite{yan2021two}. Finding ways to mitigate these problems is of great importance for advancing GNNs, and to do so, we need flexible and generalizable strategies that can easily be applied to different encoders and in both supervised and unsupervised settings.

In this work, we introduce Half-Hop, a simple yet general  augmentation for improving learning in message passing neural networks. The main idea behind our approach is to upsample the input graph: we do this through the introduction of new nodes, that we refer to as ``slow nodes'', along edges. Introducing a slow node has the effect of slowing-down the messages sent by the source node to the target node. Rather than making explicit modifications of our loss or encoder, we simply modify the input graph, making it plug-and-play and easy to use with existing models. \footnote{Code is provided at: \textcolor{cyan}{\href{https://github.com/nerdslab/halfhop}{https://github.com/nerdslab/halfhop}}.}

We apply our approach to a wide range of benchmark datasets used in supervised and self-supervised node classification tasks. Across the board, we find that our approach provides improvements to the baseline models that we tested. 
In self-supervised learning, we demonstrate impressive boosts in performance when applying our approach to state-of-the-art models for self-supervised learning (SSL) (i.e., GRACE \cite{grace}, BGRL \cite{thakoor2022large}). 
Overall, these results suggest that Half-Hop can significantly improve the performance of GNNs in a wide range of graphs, across models, losses, and tasks.

\begin{figure*}
    \centering
    \includegraphics[width=\textwidth]{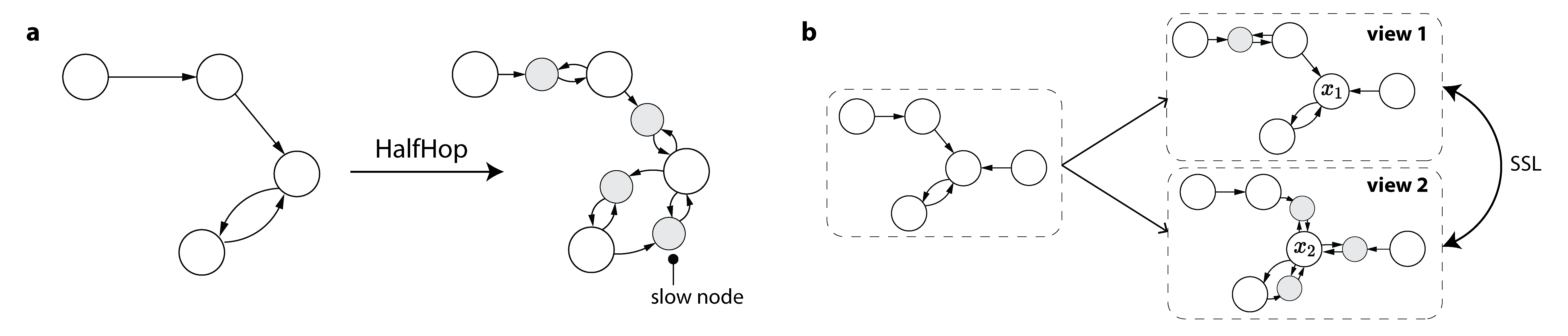}
  \vspace{-6mm}  \caption{ {\em Overview of the Half-Hop augmentation}. (a) On the left, we show an original graph and on the right, the graph after applying Half-Hop to all edges. We introduce slow nodes (gray) along each directed edge. (b) Half-Hop is used to generated diverse views for self-supervised learning methods. In the illustrated example, only the incoming edges of randomly selected nodes are half-hopped.}
    \label{fig:overview}
\end{figure*}

The contributions of this work include:
\begin{itemize}
\vspace{-3mm}
    \item {\em A graph upsampling approach that improves node classification for a range of GNNs:} In extensive experiments, we show the utility of graph upsampling in both supervised and unsupervised settings.
    \vspace{-1mm}
    \item {\em Novel graphs augmentations for SSL:} An important challenge in using SSL on graphs is the design of augmentations to create different views of a graph for contrastive learning \cite{gca}. We demonstrate that Half-Hop can be used as a standalone augmentation or in conjunction with other augmentations to improve upon state-of-the-art SSL models.
    \vspace{-1mm}
    \item {\em Plug-and-play component to improve message passing on heterophilic graphs:} Our results on heterophilic datasets show that by adding Half-Hop to a simple GCN backbone, we can achieve over 10\% boost in performance, and when combining Half-Hop with GraphSAGE \cite{hamilton2017inductive}, we achieve results that are comparable with state-of-the-art methods that use more complex architectures.
    \vspace{-1mm}
    \item {\em Study of Half-Hop and how the introduction of slow nodes helps to mitigate oversmoothing:} In both a theoretical investigation and in empirical studies, we unpack the different ways that Half-Hop impacts learning. {\color{black} In particular, we analyze the impact of Half-Hop on the dynamics of message passing and show how our approach can slow down the effects of oversmoothing.} 
\end{itemize}

\section{Method}

In computer vision, increasing the resolution of an image (zoom and crop) is one of the most widely used  augmentations for both supervised and unsupervised learning \cite{pmlr-v119-chen20j,grill2020bootstrap}. Our idea is to rescale graphs in a similar fashion. 
Just as increasing the resolution of an image requires the introduction of new  pixels, we upsample graphs by introducing new nodes along edges, and interpolating their node features based on the corresponding source and target nodes (see Figure~\ref{fig:overview}). The modified graph can then be passed into any message passing-based neural network without further modification.

\subsection{Background and Notation}
Let $G = (V, E)$ denote our graph with nodes $V$ and directed edges $E$. Each node $v_i \in V$ is associated with a set of d-dimensional features $x_{i} \in \mathbb{R}^d$. Let $e_{ij}$ denote the edge going from node $v_i$ to node $v_j$.
In the message passing scheme, nodes exchange information with their neighboring nodes, through multiple rounds of message passing. Each node $v_i$ updates its embedding by combining their ego-embedding  (the node's embedding at the previous step) and the aggregated embeddings received from their neighbors. There are multiple implementations of the message passing layer. We highlight the GCN \cite{kipf2016semi} model, for which the output at layer $\ell$ is expressed as, 
\begin{equation*}
h_i^{(\ell)} = \sigma \left(\sum_{j \in \mathcal{N}(i) \cup \{i \}} \frac{1}{\sqrt{\hat{d_j} \hat{d_i}}} \mathbf{W_\ell} h_j^{(\ell - 1)}  \right),
\end{equation*}
where $\mathcal{N}(i)$ denotes the neighbors of $v_i$,
$\mathbf{W_\ell}$ is a set of learnable shared weights used to compute messages, $\sigma$ an activation function, and $\hat{d_i} = 1 + |\mathcal{N}(i)|$.
Other GNN models include GraphSAGE which uses a seperate learnable matrix for the ego-embedding, and GAT \cite{velivckovic2017graph} which leverages attention during the aggregation step.

\subsection{Half-Hop: Graph upsampling by inserting slow nodes between adjacent nodes}
\label{sec:virtualnode}

\paragraph{Half-hopping an edge in the graph.}~We consider a directed edge $e_{ij}$ that is not a self-loop (i.e. $i\neq j$). To ``half-hop'' $e_{ij}$, we introduce a new node $\nu_k$ that splits the edge into two. The path $v_i \rightarrow v_j$ previously of length 1, is expanded: $v_i \rightarrow \nu_k \rightleftharpoons v_j$. Due to the added hop to go from $v_i$ to $v_j$, we refer to this new node as a "slow node". This modification to the graph can be expressed as follows:
\begin{equation}
    \begin{cases}
      V'= & V \cup  \{ \nu_k \} \\
      E'= & (E \setminus \{e_{ij}\}) \cup \{ e_{i\rightarrow k}, e_{j\rightarrow k}, e_{k\rightarrow j}\},  \\
      \end{cases}       
\end{equation}
Note that both the source and target nodes communicate their messages to the slow node, but the slow node only passes information in the original direction. We find that this configuration is indeed optimal compared to other connectivity motifs (see ablations in Appendix~\ref{appendix:ablations}). 

\vspace{-2mm}
\paragraph{Interpolating slow node features.}~ 
When constructing a slow node, we need to decide what features to assign to it.
A simple yet effective approach is to use linear interpolation of the source and target features. For edge $e_{i,j}$, we initialize the features of their slow node as follows:
$$
    \widetilde{x}_k = (1 - \alpha) x_j + \alpha x_i,
$$
where $x_i$ and $x_j$ are the source and target node features respectively, and $\alpha$ is some fixed scalar between 0 to 1. By adjusting $\alpha$ between 0 and 1, we can adjust the proximity of the slow node's initial features to the source or target node. We study other forms of node feature initialization in the Appendix~\ref{appendix:ablations} and show that mixing is a superior strategy when compared with random initialization or setting the features to zero. In practice, we tune the $\alpha$ parameter using a validation set; however, we find a good degree of robustness to this hyperparameter in many of our experiments. 

\vspace{-2mm}
\paragraph{Half-hopping the graph.}~Now that we have established how slow nodes are added to the graph, we can apply this operation to multiple edges in the graph at once. We can do this for all edges or for a subset of them.

In our work here, we consider a node-level sampling to apply Half-Hop randomly on a subset of edges. For each node $v_i \in V$, with probability $p$, we ``half-hop'' all of $v_i$'s incoming edges. Let $\mathcal{S}$ be the subset of nodes selected for half-hopping, $E_\mathcal{S}$ be the set of all directed edges that have target nodes in $\mathcal{S}$, where $N_s = |E_\mathcal{S}|$. The set of nodes in the new locally half-hopped graph is: $V^\prime = V \cup \{ \nu_{k}\}_{k=1}^{N_s}$.
To streamline notation, we can write a sample from this graph generative process as $(V^\prime, E^\prime) \sim hh_{\alpha}(G; p)$. When {\em all} edges in the graph are Half-Hopped (i.e. p=1), we will use an uppercase $\mathrm{HH}_{\alpha}(G)$. 
Note that in contrast to the node sampling generator, the fully half-hopped graph is a deterministic transformation.

\begin{figure*}[!t]
    \centering
    \includegraphics[width=0.98\textwidth]{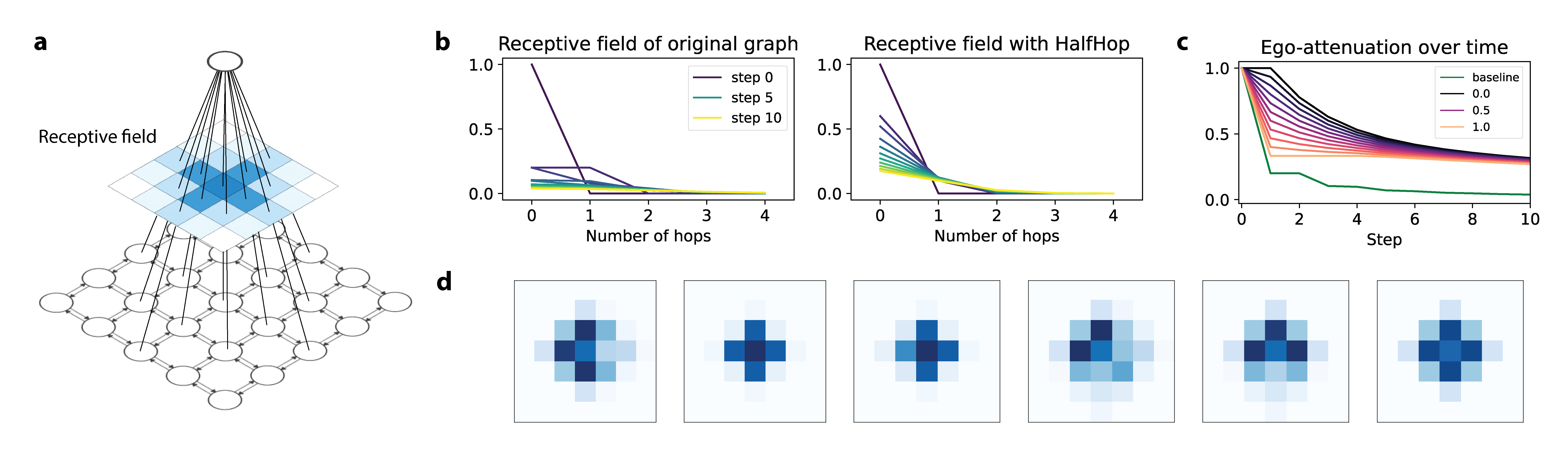}
    \vspace{-1.5em}
    \caption{\footnotesize{{\em Analyzing how Half-Hop changes the receptive field (RF) of the GNN.} (a)We consider a 2D planar graph. (b) We estimate the contribution of a node to the final embedding of the central node based on distance between the nodes (number of hops), without (b, left) and with Half-Hop (b, right). (c) The dynamics of attenuation of the ego-embedding  for different values of $\alpha$. (d) Example RFs obtained for probabilistic Half-Hop ($\alpha=0.5$, $p=0.75$) that highlight how the RF changes when different subsets of edges are half-hopped. Darker means higher contribution to the RF.}} 
    \label{fig:receptive_field}
\end{figure*}

\begin{figure*}[!h]
    \centering
    \includegraphics[width=0.98\textwidth]{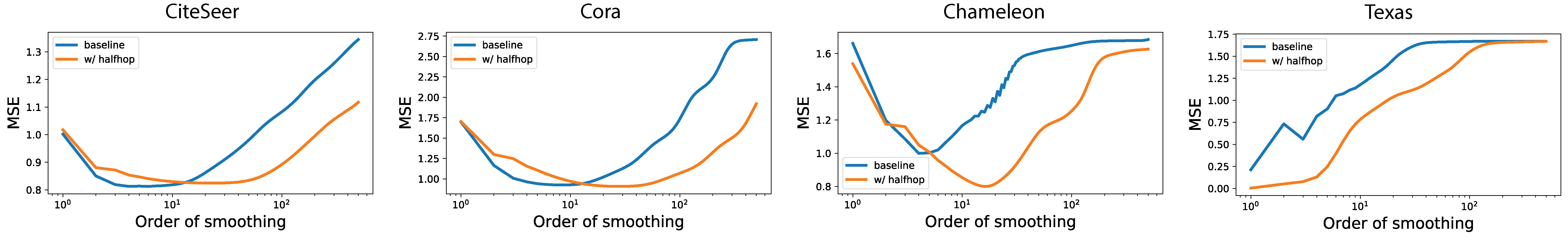}
    \caption{{\em The results of isotropic diffusion with and without Half-Hop.} Mean Squared Error (MSE) of linear ridge regression on the smoothed features after a number of rounds of message passing, described as order of smoothing (reported in log scale). From left to right, we illustarate it for CiteSeer, Cora, Chameleon and Texas. 
    }
    \label{fig:oversmoothing}
\end{figure*}

\subsection{Combining Half-Hop with any MPNN}
After applying Half-Hop, we obtain  a modified graph that can be passed into a message passing neural network. 
We treat original and slow nodes the same when it comes to MP, both nodes receive and send updates according to the same rules.
We treat the slow nodes as intermediary nodes that are discarded of at the end of MP, and any downstream operations such as loss computation or inclusion in further neural network encoders work only over the embeddings of the original nodes.

After training our model with Half-Hop, at inference time, we have the choice of whether to use the original graph $G$ or the Half-Hopped graph $\mathrm{HH}_{\alpha}(G)$ as input. If we choose to not augment the graph at inference time, we will simply benefit from an improved model generalization, enabled by the use of Half-Hop as an augmentation during training (as we show in Section 4.4). If we do augment the graph, we would also take advantage of the improved MP enabled by Half-Hop (as we demonstrate in Section 3.2). 

\subsection{Using Half-Hop to generate views for Self-supervised learning}
\label{sec:ssl}
We also consider the use of Half-Hop for self-supervised representation learning.
Self-supervised learning methods use augmentations to  generate different views of a graph and then learn a representation space in which these views are close to each other \cite{zhu2020beyond,thakoor2022large}. 
With Half-Hop, we can create views where node $i$ might be directly connected to its one-hop neighbors in one view, but half-hopped in the other. This generates a heterogeneous zooming and cropping into the neighborhood of different nodes. We expand on this in Section 3.2, where we show how Half-Hop alters the receptive field of the GNN.

Thus, we can generate two views $(\widetilde{G}_1, \widetilde{G}_2)$ of the graph to use as inputs to contrastive learning:
$$\widetilde{G}_1 \sim hh_{\alpha} ( G; p_1), \widetilde{G}_2 \sim hh_{\alpha} ( G; p_2)$$
where we parameterize the sampling procedure for each view with node-sampling probabilities $p_1$ and $p_2$. 
As explained in the previous section, when evaluating the contrastive loss, we only use the original nodes as positive/negative examples.

\section{Understanding Half-Hop}

In this section, we investigate Half-Hop from two angles. In Section~\ref{sec:RF}, we study how Half-Hop alters the receptive field of the GNN model. In Section~\ref{sec:theory}, we study the effect of Half-Hop on message passing from a spectral perspective.
In both theory and experiments, we show that our approach slows down the smoothing process.

\subsection{Reshaping the node's receptive field with Half-Hop}
\label{sec:RF}

The receptive field (RF) of a model represents the parts of the input graph that have the most significant impact on the final embedding of a particular node. In MPNNs, the representation of a node is typically influenced by its neighboring nodes.
When Half-Hop is applied, 1-hop neighbors become 2-hop neighbors and the receptive field is thus reduced since messages take longer to propagate as they are routed through slow nodes. Thus, we wanted to understand how the RF is being shaped over rounds of message passing. To do this, we consider a simple 2D planar graph (Figure~\ref{fig:receptive_field}a), and a simplified GNN that is equivalent to applying multiple rounds of mean aggregation without any learned weights. If we note ${H}^{(0)}$ the initial feature matrix of the nodes, and ${H}^{(k)}$ the output after $k$ message passing steps, then we have ${H}^{(k)} = {L}^k {H}^{(0)}$ with $L=D^{-1}A$, $A$ denotes the adjacency matrix and $D$ the diagonal degree matrix. In other words, the output embedding of a node can be expressed as a weighted combination of all nodes in the graph.

In Figure~\ref{fig:receptive_field}b, we visualize, for a given central node in this simple 2D graph, the contribution (weight) of any $k$-hop neighbor to the central node's final embedding. In the original graph (Fig.~\ref{fig:receptive_field}b, left), the weights of messages from nearest neighbors are quickly attenuated. Even after one step, the uniform weights between 1-hop neighbors and self-loops flattens the RF immediately from the central node. This smoothing process can often take place extremely quickly, as demonstrated in this example. When we apply Half-Hop (Fig.~\ref{fig:receptive_field}b, right), we show that the receptive field is altered: a more graceful decrease in the weight of neighbors is observed, with a higher amount of weight placed on the node itself.  
In Figure~\ref{fig:receptive_field}c, we plot the self-weight (y-axis) as a function of the message passing step (x-axis), and show that Half-Hop allows us to preserve the ego-embedding for nodes for longer. Even as we converge to large values of $\alpha$, Half-Hop preserves a stable value for the self-weight due to the symmetry in mixing across source and target nodes. At the same time, most values of $\alpha$ have similar RFs when considering later stages of MP. 

In self-supervised learning, we use Half-Hop to generate multiple views in which the same node can have different receptive fields. In Figure~\ref{fig:receptive_field}d, we highlight a few examples of receptive fields of the same central node when Half-Hop is applied with $\alpha=0.5$ and $p=0.75$ on the grid graph. Because we sample a subset of the nodes for half-hopping, we can generate diverse configurations. We find that the receptive field can be narrow, broad or a hybrid of both based on which nodes were half-hopped. This augmentation is reminiscent to zooming and cropping in the vision domain, in that we aggregate information at different scales.

\subsection{Half-Hop shapes the dynamics of message passing}

{\color{black} 
In simple settings, Keriven (\citeyear{keriven2022not}) show that the dynamics of message passing can be characterized, and that the underlying feature covariance and topology play an important role. Thus, we sought to utilize this framework to provide insight into how Half-Hop may alter message passing dynamics and induce helpful forms of regularization into learning.
}

\vspace{-2mm}
\subsubsection{Diffusion on Real-World Datasets} 
To first simulate the effect of diffusion on the Half-Hop graph vs. the original graph  for four different real-world datasets (CiteSeer, Cora, Chameleon, Texas), we use a simplified linear GNN with multiple rounds of message passing (no learned weights), followed by a final linear layer that is trained on a subset of nodes and tested on held-out nodes. In Figure~\ref{fig:oversmoothing}, we report, for different datasets, the test mean squared error (MSE) for the node regression task as a function of the number of message passing steps. We note that CiteSeer/Cora are considered homophilic, while Chameleon/Texas are heterophilic.

When we compare the generalization dynamics for Half-Hop vs. the baseline, we find very different behavior for homophilic (left) vs. heterophilic graphs (right). In the homophilic graphs, we achieve overall similar rates of test error with the baseline and Half-Hop but we see the basin of low error solutions is widened (suggesting more stability) and the point where oversmoothing kicks in (MSE starts to go up again) is increased. Naturally, we would expect the factor to be at least twice as large, since Half-Hop doubles the diameter of the graph, but we find that the transition point is even greater than what would be predicted by this factor.

When we examine heterophilic graphs (Chameleon, Texas), we observe that Half-Hop achieves a significantly lower risk than the baseline. In the case of Chameleon, Half-Hop improves the descent and overall test risk significantly, this can be explained by the fact that heterophilic graphs suffer more from the oversmoothing, since we would be aggregating features of nodes that do not belong to the same class.  This demonstrates the useful inductive bias in the Half-Hop model even without learning weights for message passing. In Texas, any rounds of mean aggregation (with learning of weights) hurt the test risk; however, Half-Hop achieves a much lower risk at the first few steps of message passing and also stabilizes the risk far longer. In Appendix~\ref{app:latentviz}, we visualize how the embeddings changes over multiple rounds of MP, and show how Half-Hop also slows down the collapse of the latent space.

\begin{table*}[!ht]
\centering
\caption{{\em  Increase in supervised performance when using Half-Hop.  \label{tab:supervised}} 
The average and standard deviation of accuracy is computed over 20 random splits and model initializations. The absolute improvement ($\Delta$) is also reported.}
\vspace{3mm}
\resizebox{0.67\textwidth}{!}{
\begin{tabular}{lcccc}
\hline
 & Am. Comp. & Am. Photos   & Co.CS      & WikiCS \\
 \hline
GCN  & 90.22 $\pm$ 0.60 & 93.59 $\pm$ 0.42 & 94.06 $\pm$ 0.16 & 81.93 $\pm$ 0.42 \\
HH-GCN  & 90.92 $\pm$ 0.35 & 94.52 $\pm$ 0.22 & 94.71 $\pm$ 0.16 &  82.57 $\pm$ 0.36 \\   
$\Delta$ & \textbf{+0.70} ($\uparrow$) & \textbf{+0.93} ($\uparrow$) & \textbf{+0.65} ($\uparrow$) & \textbf{+0.64} ($\uparrow$) \\
\hline
GraphSAGE & 84.79 $\pm$ 1.08 & 95.03 $\pm$ 0.33 & 95.11 $\pm$ 0.10 & 83.67 $\pm$ 0.45\\            
HH-GraphSAGE & 86.60 $\pm$ 0.49 & 94.55 $\pm$ 0.41 & 95.13 $\pm$ 0.21 & 82.81 $\pm$ 0.32\\
$\Delta$ & \textbf{+1.81} ($\uparrow$) & \textbf{-0.48} ($\downarrow$) & \textbf{+0.02} ($\uparrow$) &  \textbf{-0.86} ($\downarrow$)  \\
\hline
\vspace{-3mm}
\end{tabular}}
\end{table*}

\begin{table*}[t!]
\centering
\vspace{-2mm}
\caption{{\em  Results on heterophilic graphs.} We report the test accuracy across many heterophilic graph benchmark datasets, and highlight the absolute improvement ($\Delta$) in classification accuracy when the model is augmented with Half-Hop. {\color{black} The ``$\dag$'' results are obtained from \cite{yan2021two}}.\label{tab:hetero}} 
\vspace{3mm}
\resizebox{0.89\textwidth}{!}{
\begin{tabular}{lcccccc}
\hline
& Texas & Wisconsin & Actor  & Squirrel & Chameleon & Cornell  \\
 Hom level & 0.11 & 0.21 & 0.22 & 0.22 & 0.23 & 0.30 \\
\#Nodes & 183 & 251 & 7,600 & 5,201 & 2,277 & 183 \\
\#Edges & 295 & 466 & 26,752 & 198,493 & 31,421 & 280 \\
\#Classes & 5 & 5 & 5 & 5 & 5 & 5 \\
 \hline
GCN$^\dag$  & 55.14 $\pm$ 5.16 & 51.76 $\pm$ 3.06  & 27.32 $\pm$ 1.10 & 31.52 $\pm$ 0.71 & 38.44 $\pm$ 1.92 & 60.54 $\pm$ 5.30 \\
HH-GCN & 71.89 $\pm$ 3.46 & 79.80 $\pm$ 4.30 &  35.12 $\pm$ 1.06 & 47.19 $\pm$ 1.21 & 60.24 $\pm$ 1.93 & 63.24 $\pm$ 5.43 \\
$\Delta$ & \textbf{+16.75} ($\uparrow$) &  \textbf{+19.04} ($\uparrow$) & \textbf{+7.80} ($\uparrow$) & \textbf{+15.67} ($\uparrow$) & \textbf{+21.80} ($\uparrow$) & \textbf{+2.70} ($\uparrow$) \\
\hline
GAT$^\dag$  & 52.16 $\pm$ 6.63 & 49.41 $\pm$ 4.09 & 27.44 $\pm$ 0.89 & 36.77 $\pm$ 1.68 & 48.36 $\pm$ 1.58 & 61.89 $\pm$ 5.05 \\
 HH-GAT & 80.54 $\pm$ 4.80 & 83.53 $\pm$ 3.84 & 36.70 $\pm$ 0.92 & 46.35 $\pm$ 1.86 & 61.12 $\pm$ 1.83 & 72.70 $\pm$ 4.26\\
$\Delta$ & \textbf{+28.38} ($\uparrow$) &  \textbf{+34.12} ($\uparrow$) & \textbf{+9.26} ($\uparrow$) & \textbf{+9.58} ($\uparrow$) &  \textbf{+12.75} ($\uparrow$) & \textbf{+10.81} ($\uparrow$) \\
\hline
 GraphSAGE$^\dag$  & 82.43 $\pm$ 6.14 & 81.18 $\pm$ 5.56 & 34.23 $\pm$ 0.99 & 41.61 $\pm$ 0.74 & 58.73 $\pm$ 1.68 & 75.95 $\pm$ 5.01 \\
HH-GraphSAGE  & 85.95 $\pm$ 6.42 & 85.88 $\pm$ 3.99 & 36.82 $\pm$ 0.77 & 45.25 $\pm$ 1.52 & 62.98 $\pm$ 3.35 & 74.60  $\pm$ 6.06 \\
$\Delta$ & \textbf{+3.51} ($\uparrow$) &  \textbf{+4.70} ($\uparrow$)& \textbf{+2.59} ($\uparrow$) & \textbf{+3.64} ($\uparrow$) & \textbf{+4.25} ($\uparrow$) & \textbf{-1.35} ($\downarrow$) \\
\hline 
MixHop$^\dag$  & 77.84 $\pm$ 7.73 & 75.88 $\pm$ 4.90 & 32.22 $\pm$ 2.34 & 43.80 $\pm$ 1.48 & 60.50 $\pm$ 2.53 & 73.51 $\pm$ 6.34\\
GGCN$^\dag$  & 84.86 $\pm$ 4.55 & 86.86 $\pm$ 3.29 & 37.54 $\pm$ 1.56 & 55.17 $\pm$ 1.58 & 71.14 $\pm$ 1.84 & 85.68 $\pm$ 6.63 \\
H$_2$GCN$^\dag$  & 84.86 $\pm$ 7.23 & 87.65 $\pm$ 4.98 & 35.70 $\pm$ 1.00 & 36.48 $\pm$ 1.86 & 60.11 $\pm$ 2.15 & 82.70 $\pm$ 5.28 \\
\hline
MLP$^\dag$   & 80.81 $\pm$ 4.75 & 85.29 $\pm$ 3.31 & 36.63 $\pm$ 0.70 & 28.77 $\pm$ 1.56 & 46.21 $\pm$ 2.99 & 81.89 $\pm$ 6.40 \\
\hline
\end{tabular}
}
\end{table*}

\subsubsection{Analyzing the effect of Half-Hop on generalization}
\label{sec:theory}
{\color{black} 
Our simulations suggest that Half-Hop does indeed alter the dynamics of message passing. Thus, we wanted to dig deeper and develop a result that allows us to compare the dynamics of message passing with and without Half-Hop. Throughout, we follow the assumptions and model described in \cite{keriven2022not}. We point the reader to their work for the full analysis and Appendix~\ref{appendix:analysis} for further details and proofs. 
}

{\color{black}
{\em A) Preliminaries and assumptions:}~

{\bf Graph Model.}~We adopt the latent space random graph model, where 
we assume that the observed node features $x_i=M^\top z_i$ are projections of some underlying latent features $z_i$, where  $M$ denotes an unknown projection matrix; We assume the latent features $z_i \sim \mathcal{N}(0,\Sigma)$, with covariance $\Sigma$. The edge weights are determined as a function of the latent variables by $W_{i,j}=\varepsilon + \exp(-\frac{1}{2}\|z_i-z_j\|_2^2)$, where $\varepsilon$ is an unknown offset. The node labels, $y_i=z_i^\top\beta^{\ast}$, are linear functions of the latent variable $z_i$ with unknown coefficients $\beta^{\ast}$.

{\bf Objective.}~We consider a semi-supervised ridge regression task where the goal is to estimate $\beta^{\ast}$ and use it to predict the labels for nodes in the test set. 
We use MSE to write the test risk as  $\mathcal{R}^{(k)} = \| Y_{te} - \widehat{Y}_{te} \|^2$,  where $Y_{te}$ are the stacked labels for $n_{te}$ unlabeled nodes in the test set, $\widehat{Y}_{te} = X_{te}^{(k)} \widehat{\beta}$ are the estimated labels  after k steps of message passing (diffusion), $X_{te}^{(k)}$ are the corresponding node features, and $\widehat{\beta}$ is estimated on the training set.

{\em B) How the spectrum impacts the dynamics of message passing:}~
To present our main result, we first need to define the following function which we will use to bound the risk.

{\bf Definition 1. \cite{keriven2022not}}~For a symmetric positive semi-definite matrix $S \in R^{d \times d}$, we define the function,
\begin{equation*}
R_{\text {reg. }}(S) \stackrel{\textrm { def. }}{=}\left(\Sigma^{\frac{1}{2}} \beta^{\star}\right)^{\top} K \left(\Sigma^{\frac{1}{2}} \beta^{*}\right) \in \mathbb{R}_{+}
\vspace{-2mm}
\end{equation*}
\textrm{where}~ $K = \left(\mathrm{I}-S^{\frac{1}{2}} M\left(\gamma \mathrm{I}+M^{\top} S M\right)^{-1} M^{\top} S^{\frac{1}{2}}\right)^2$, $\Sigma$ is the latent model covariance, $M$ is the  projection matrix,  $\beta^*$ are the true model parameters, and $\gamma$ is the ridge penalty in our least-squares estimator.

Following Keriven (\citeyear{keriven2022not}), one can show that the risk without message passing (MP) can be approximated by 
$\mathcal{R}^{(0)}\simeq R_{\text {reg.}}\left(\Sigma\right)$, and the risk after $k$ rounds of MP can similarly be approximated as  $\mathcal{R}^{(k)}\simeq R_{\text {reg.}}\left(A^{2k}\Sigma\right),
$  where $A= (\mathrm{I} + \Sigma^{-1})^{-1}$. In this case, we can interpret $A$ as a smoothing operator that is applied to the original spectrum, and interpret $\Sigma^{(k)} = A^{2k}\Sigma$ as the modified covariance after k rounds of MP.

{\em C) Main Result:}~ 
Our goal is to derive a similar result to understand how Half-Hop: (i) impacts the feature covariance of the embeddings, (ii) changes the rate of smoothing as we go deeper and run more rounds of message passing. To do this, we want to derive an approximation of the risk in the form $R_{\text {reg. }}(\Sigma^{(\mathrm{HH},k)})$ where $\Sigma^{(\mathrm{HH},k)}$ is the approximated covariance of the node features after message passing with Half-Hop. 
We state our main result below and defer the proof to  Appendix~\ref{appendix:analysis}.

\textbf{Theorem 1. Message Passing Dynamics of Half-Hop.}  
After
$k\in\{1,3,5,\dots\}$ rounds of message passing, the risk obtained with
\textit {Half-Hop} can be approximated as:
\vspace{-2mm}
\begin{align*}
\mathcal{R}^{(\mathrm{HH},k)}&\simeq R_{\text {reg. }}\left(\frac{1}{2}A^{k-1}\left(\textrm{I}+\left((1-\alpha) \textrm{I}+\alpha A\right)^2\right)\Sigma\right).
\vspace{-1mm}
\end{align*}}
{\color{black}
The first term $A^{k-1}\Sigma$ smooths the covariance at a rate of $k-1$, which is roughly half the rate of smoothing without Half-Hop ($2k$). \cite{keriven2022not} links the rapid decay of small eigenvalues in particular to the over-smoothing phenomena. Half-Hop thus ensures that small eigenvalues decay at a slower rate, and thus delays the point at which the diffusion stops being beneficial.

The second term in our augmented covariance, $ ((1-\alpha) \textrm{I}+\alpha A )^2 $,  reveals the dependence on our mixing parameter $\alpha$. In particular, we observe a uniform boost of the covariance spectrum coming from the $(1-\alpha)\text{I}$; for small values of $\alpha$, this term amplifies self-loops and the small eigenvalues.
}

\begin{table*}[!th]
\centering
\caption{{\em  BGRL with different augmentations.} Performance reported in terms of classification accuracy along with standard deviation. All experiments are performed over 20 random dataset splits and model initializations. At test time, the original graph is used. OOM indicates out-of-memory on a 48GB Nvidia A40 GPU.  The ``$\dag$'' results are obtained from \cite{thakoor2022large}. \label{tab:ssl_augmentations}}
\vspace{3mm}
\resizebox{\textwidth}{!}{
\begin{tabular}{l|l|ccccc}
\hline
& Augmentation                       & Am. Comp.        & Am. Photos       & Co.CS            & Co.Phy           & Wiki-CS   \\
 \hline
\multirow{5}{*}{BGRL}  & None        & 87.12 $\pm$ 0.30 & 91.18 $\pm$ 0.38 & 91.85 $\pm$ 0.25 & 94.65 $\pm$ 0.11 & 78.69 $\pm$ 0.18 \\
  & FeatDrop + EdgeDrop              & 90.34 $\pm$ 0.19 & 93.17 $\pm$ 0.30 & 93.31 $\pm$ 0.13 & {\bf 95.73 $\pm$ 0.05} & 79.98 $\pm$ 0.10 \\
  & GCA$^\dag$                              & 90.39 $\pm$ 0.22 & 93.15 $\pm$ 0.37 & 93.34 $\pm$ 0.13 &  95.62 $\pm$ 0.09 & -- \\ 
  & Half-Hop                          & 90.47 $\pm$ 0.25 & 93.18 $\pm$ 0.26 & 92.92 $\pm$ 0.11 & 95.69 $\pm$ 0.21 & 79.83 $\pm$ 0.53\\
  & FeatDrop + EdgeDrop + Half-Hop    & {\bf 91.02 $\pm$ 0.27} & 93.88 $\pm$ 0.19 & {\bf 93.61 $\pm$ 0.13} & {\bf 95.75 $\pm$ 0.13} & {\bf 80.76 $\pm$ 0.71}\\
\hline
\multirow{5}{*}{GRACE}  & None       & 77.85 $\pm$ 0.96 & 88.47 $\pm$ 0.67 & 90.04 $\pm$ 0.36 & OOM              & 70.61 $\pm$ 0.95 \\
 & FeatDrop + EdgeDrop               & 89.53 $\pm$ 0.35 & 92.78 $\pm$ 0.45 & 91.12 $\pm$ 0.20 & OOM              & 80.14 $\pm$ 0.48\\
 & GCA$^\dag$                              & 87.85 $\pm$ 0.31 & 92.49 $\pm$ 0.09 & 93.10 $\pm$ 0.01 & OOM              & -- \\
 & Half-Hop                           & 90.43 $\pm$ 0.28 & 93.58 $\pm$ 0.18 & 92.29 $\pm$ 0.12 & OOM              & 79.86 $\pm$ 0.41 \\
 & FeatDrop + EdgeDrop + Half-Hop     & {\bf 91.11 $\pm$ 0.18} & {\bf 94.21 $\pm$ 0.26} & {\bf 93.59 $\pm$ 0.16} & OOM              & {\bf 80.77 $\pm$ 0.40}\\ 
\hline
\end{tabular}}
\end{table*}

\begin{table*}[!ht]
\centering
\caption{{\em  Increase in performance when using Half-Hop with different SSL frameworks.} Encoders 2-GCN and 3-GCN represent a 2 layer and a 3 layer GCN respectively. Input (test) denotes the graph supplied at test time, where we can choose to use the original graph G or the augmented graph HH(G). Performance is reported in terms of classification accuracy along with standard deviations. All experiments are performed over 20 random dataset splits and model initializations. \label{table:SSL}}
\vspace{3mm}
\resizebox{\textwidth}{!}{
\begin{tabular}{lcc|ccccc}
\hline
 & input (test) & encoder & Am. Comp. & Am. Photos   & Co.CS       & Co.Phy       & WikiCS \\
 \hline
GRACE & G & 2-GCN & 89.53 $\pm$ 0.35 & 92.78 $\pm$ 0.45 & 91.12 $\pm$ 0.20 & OOM & 80.14 $\pm$ 0.48\\
HH-GRACE & G & 2-GCN & \bf 91.11 $\pm$ \bf 0.18 & 94.21 $\pm$ 0.26 & 93.59 $\pm$ 0.16 & OOM & 79.77 $\pm$  0.40 \\
HH-GRACE & HH(G) & 2-GCN & 90.65 $\pm$ 0.19 & \textbf{94.89 $\pm$ 0.23} & \textbf{94.76 $\pm$ 0.14}  & OOM & 80.15 $\pm$ 0.16\\
\hline
BGRL & G & 2-GCN &  90.34 $\pm$ 0.19 & 93.17 $\pm$ 0.30 & 93.31 $\pm$ 0.13  & 95.73 $\pm$ 0.05 & 79.98 $\pm$ 0.10 \\
HH-BGRL & G & 2-GCN & 91.02 $\pm$ 0.27 & 93.88 $\pm$ 0.19 & 93.61 $\pm$ 0.13 & 95.75 $\pm$ 0.13 & 80.76 $\pm$ 0.71\\
HH-BGRL  & HH(G)& 2-GCN & 90.94 $\pm$ 0.19 & 94.50 $\pm$ 0.35 & \bf 94.74 $\pm$ \bf 0.15 & \bf \bf 96.13 $\pm$ \bf 0.10 & 80.37 $\pm$ 0.62 \\
\hline
BGRL & G & 3-GCN & 90.04 $\pm$ 0.23 & 92.59 $\pm$ 0.34 & 92.42 $\pm$ 0.17 & 95.32 $\pm$ 0.51 & 78.22 $\pm$ 0.77 \\
HH-BGRL & G & 3-GCN & 90.53 $\pm$ 0.27 & 93.09 $\pm$ 0.16 & 92.58 $\pm$ 0.20 & 95.45 $\pm$ 0.09 & 79.76 $\pm$ 0.61\\
HH-BGRL & HH(G)& 3-GCN & \bf 91.10 $\pm$ \bf 0.21 & 94.34 $\pm$ 0.25 & {\bf 94.76 $\pm$ 0.12} & \bf 96.10 $\pm$ \bf 0.09 & \bf 81.11 $\pm$ \bf 0.48 \\
\hline
\end{tabular}}
\vspace{-2mm}
\end{table*}

\section{Experimental Results}
In this section, we conduct a comprehensive empirical study of the effectiveness of Half-Hop on a wide range of datasets, models and learning paradigms.

\subsection{Experimental Setup}

Throughout, we test our approach using three of the most widely used graph models: the Graph Convolutional Network (GCN), GraphSAGE, and Graph Attention Network (GAT). In all of our experiments, we follow the same experimental setup as previous work. In the supervised experiments, we follow \cite{math10081262} in splitting the data into a development set and a test set. The hyperparameter tuning is performed using the development set only, and the accuracy of the best model on the development set is reported on the test set. For heterophilic datasets, we use the splits provided by Pei et al. (\citeyear{Pei2020Geom-GCN}), and also follow the same hyperparameter search protocol. 
For self-supervised benchmarks, we use the standard hyperparameters provided for each model and dataset \cite{grace,thakoor2022large}. We provide more details in Appendix \ref{appendix:supex}.

\subsection{Supervised node classification benchmarks}

In our first set of experiments, we use a set of real-world benchmark datasets -- Wiki-CS \cite{mernyei2020wiki}, Amazon-Computers, Amazon-Photo \cite{amazon}, and Coauthor datasets \cite{coauthor}. 
We train both GCN and GraphSAGE models with and without Half-Hop, and report the results in Table~\ref{tab:supervised}, where we note the Half-Hop variants HH-GCN and HH-GraphSAGE respectively. 
Our results show that Half-Hop provides a good boost in performance across the datasets for the GCN backbone, while GraphSAGE has more variability. HH-GraphSAGE on Amazon Computers is the most significant, where we observe a 1.81\% improvement with Half-Hop over the baseline. These results provide evidence that Half-Hop can improve learning using only a simple augmentation of the graph.

\begin{figure*}[!t]
    \centering
    \includegraphics[width=\textwidth]{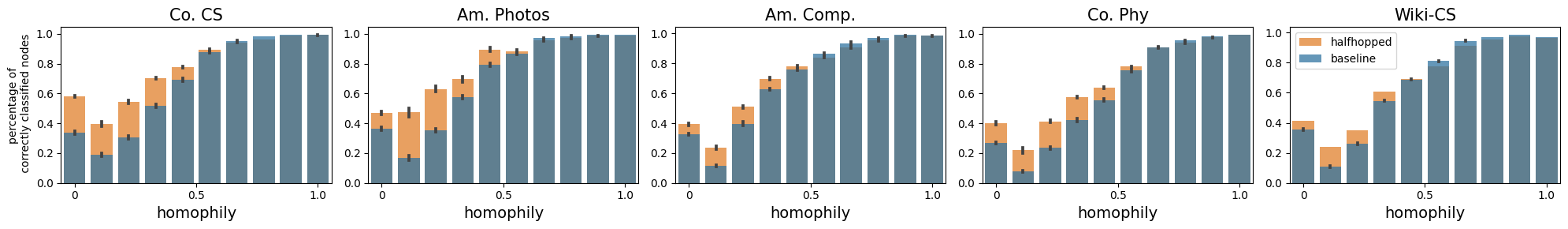}
    \vspace{-6mm}
    \caption{{\em An analysis of the accuracy on SSL baselines across different levels of heterophily}, when using HH-BGRL vs. the BGRL baseline. The orange represents the relative improvement obtained by using the Half-Hop augmentation at inference time compared to the baseline (blue). Nodes are ordered by their homophily with the most heterophilic nodes to the left. On all datasets tested, the boosts appear to be most significant on heterophilic nodes.}
    \label{fig:supervised}
\end{figure*}

\subsection{Heterophilic benchmarks}
In Table~\ref{tab:hetero}, we study the performance of Half-Hop on a number of common real-world heterophilic benchmarks, including: Texas, Wisconsin, Actor, Squirrel, Chameleon and Cornell \cite{heterophily}. 
On these datasets, we show improvements across the board when we add Half-Hop to  GCN, GraphSAGE and GAT. With both HH-GCN and HH-GAT, we see improvement of more than 10-20\% on many of the datasets. To place these improvements in the context of more sophisticated methods for heterophilic graphs, we also compare with: (i) MixHop \cite{abu2019mixhop},  (ii) GGCN \cite{yan2021two}, and (iii) $H_2$GCN \cite{zhu2020beyond}. See Table~\ref{fig:GNNcomparisons} in the Appendix for a discussion of the assumptions and components underlying these different approaches.

We find that Half-Hop boosts performance most significantly for the GCN and GAT models, with modest (2.7\%) improvements with GCN  on Cornell and a huge (21.8\%) improvement on Chameleon. For GAT, we see even larger improvements on Texas where we get a 28.38\% boost with Half-Hop.
The GraphSAGE encoder achieves the best performance out of the three models, and HH-GraphSAGE model reaches a performance that is comparable to the other competitor approaches that use more complex model components. We find that our approach provides an impressive gain for simple architectures that rivals with these other models without the need to define complex heuristics or specialized architectures. 

\subsection{Self-supervised learning benchmarks}

Graph representation learning methods rely on augmentations that are based on random transformations of the input. Thus, we can test the utility of Half-Hop for creating views for self-supervised learning and also as an add-on with existing augmentations, notably FeatDrop and EdgeDrop \cite{grace}. To do this, we combined Half-Hop with two state-of-the-art methods for self-supervised learning, BGRL \cite{thakoor2022large} and GRACE \cite{grace}. 
In this case, we report two sets of results: (i) using Half-Hop as an augmentation during training and then applying the model to the original graph $G$ at inference, (ii) applying Half-Hop during both training and inference.

{\bf Using Half-Hop to generate views.}~
First, we examined how Half-Hop could be combined with the existing augmentations used in BGRL and GRACE, and how well it performs as an augmentation on its own. We report the results in Table~\ref{tab:ssl_augmentations}, where we show that by combining standard augmentations used in BGRL and GRACE with Half-Hop, we get a nice boost over the BGRL baseline and even more impressive improvement for the GRACE model (2\% on Am.~Photos and on Co.~CS). Interestingly, when we use Half-Hop as the standalone augmentation, we perform comparably to models trained with FeatDrop and EdgeDrop or their adaptive variant GCA \cite{gca}.

{\bf Using Half-Hop at test time.}~As we observed in the supervised case, using Half-Hop at test time leads to improved message passing. We test $\mathrm{HH}$-GRACE and $\mathrm{HH}$-BGRL with both the original graph $\mathrm{G}$ and the half-hopped graph $\mathrm{HH(G)}$. Our results in Table~\ref{table:SSL} provide compelling evidence that Half-Hop improves self-supervised learning, in some cases by a significant margin. With HH-GRACE, we see an improvement of more than 2\% on Amazon-Photos and Coauthor-CS. We also see similar improvements in HH-BGRL over the baseline which uses a 2-layer GCN.

When we make the GCN in BGRL deeper (3 layers), we find that the performance degrades by an average of $0.8$\%. When we do the same with HH-BGRL, we find that increase the depth leads to even higher performance. In particular, we find a significant enhancement for HH-BGRL on the dataset with the lowest homophily ($0.66$), WikiCS, with added depth. To better understand the sources of these improvements, we breakdown the node-level accuracies by homophily and show that more heterophilic nodes are classified correctly (Figure~\ref{fig:supervised}).

\section{Related Work}

\subsection{Graph data augmentations for regularization}
\vspace{-1mm}
Data augmentation is widely used in graph learning to improve the robustness and generalization capabilities of models. Thus, this has spurred a lot of interest in designing augmentations for graph-structured data \cite{gca}. 

{\bf Feature perturbations.}
The most common feature-based augmentation is feature masking or feature dropout \cite{you2020graph, dgi}, which involves randomly setting a subset of a node's features to zero. Other approaches like FLAG \cite{kong2022robust} and LA \cite{liu2022local} use generative modeling to introduce gradient-based adversarial perturbations to the node's features. Mixup for graphs has also been proposed but usually requires a particular architecture \cite{graphmix}, or a sub-graph sampling strategy \cite{wang2021mixup}. 

{\bf Edge Perturbation.}
Adding and removing edges can also be used to perturb the connectivity of a node. In DropEdge \cite{Rong2020DropEdge,NEURIPS2020_fb4c835f,dgi}, each edge has a given probability of being removed. In GCA \cite{gca}, an edge is more or less likely to be removed based on the connectivity of its target node. GCC \cite{10.1145/3394486.3403168} uses random walks to sample ego-networks around a central node.  Approaches that add edges, on the other hand, typically require more guidance, like GAug \cite{DBLP:journals/corr/abs-2006-06830} which uses neural edge predictors to infer the probability of an edge.

{\bf k-Hop augmentations.}
Adding edges can also be used to connect more distant neighbors (k-hops away). This is typically achieved using a diffusion process \cite{8920070, pmlr-v119-hassani20a}. With this type of augmentation, we are effectively expanding the receptive field of the GNN, and are able to replicate the zoom-out operation that we know in images. On the other hand, our proposed augmentation, reduces the receptive field and allows a zoom-in operation in that sense.

\vspace{-3mm}
\subsection{Graph manipulation at inference time}
\vspace{-1mm}
When graphs are sparsely connected, highly heterophilic or have bottlenecks, graph rewiring techniques are used to correct the connectivity of nodes. SDRF \cite{ricciflow} adds edges based on Ricci curvature, EGP \cite{deac2022expander} generates edges from Cayley graphs, while DIGL \cite{gasteiger_diffusion_2019} uses a diffusion process to add edges. NeuralSparse \cite{pmlr-v119-zheng20d} and GGCN \cite{yan2021two} are supervised methods that use labels to learn to remove task-irrelevant edges that typically cause oversmoothing in heterophilic neighborhoods. 
\cite{ding2018semi} uses a generative modeling framework to create a small number of nodes that connect different subgraphs.

\vspace{-2mm}
\subsection{Self-supervised and contrastive learning methods}
\vspace{-1mm}
Many state-of-the-art approaches for self-supervised learning (SSL) on graphs use augmentations to create different views (i.e., positive examples) and then encourage the representations of both views to be close in the latent space. For instance, GRACE \cite{grace} uses a contrastive loss to encourage positive examples (a new graph with dropped edges and node features) to become closer to one another, while considering all other nodes to be far away. BGRL instead uses a score-based approach that doesn't explicitly incorporate negative examples into the loss \cite{thakoor2022large}.
Most of these approaches use relatively simple augmentations, like node and edge dropout, to create views for learning. However, adaptive graph augmentations like those proposed in GCA \cite{gca} that use topology-level and node-attribute-level augmentations like  ‘node centrality’ measures can also be used to determine edges to mask.
\vspace{-1mm}

\section{Discussion}

In this work, we introduced Half-Hop, a novel graph augmentation for message passing neural networks. Half-Hop provides a simple and yet effective approach for improving message passing: it operates by slowing down messages through the introduction of ``slow nodes'' that delay communication. We show the promise of this approach in a wide range of applications and across diverse sets of graph benchmarks.

Our experiments on 11 real-world datasets, spanning both supervised and self-supervised settings, highlight the robustness and wide applicability of Half-Hop. In heterophilic settings, we observe impressive boosts in performance when Half-Hop is added to simple encoders like the GCN, making them on par with specialized models that are adapted to heterophilic conditions. In self-supervised learning, our model also improves over SOTA graph representation learning methods, where we show it can be used as a standalone augmentation or can be coupled with existing augmentations. Overall, we show that Half-Hop is a practical, plug-and-play augmentation that integrates seamlessly into existing workflows.

Our theoretical analysis helped identify connections between the Half-Hop augmentation and how it impacts the dynamics of message passing. In particular, we provided an approach for linking the spectral effects of augmentations to the efficacy and robustness of graph learning. In the future, we anticipate that leveraging the growing line of work \cite{lin2022good} for studying the effects of data augmentations on model generalization, can provide an avenue for providing a more in depth analysis of how graph augmentations like Half-Hop impact generalization, and help to devise new augmentations. 

The choice of augmentations in SSL remains critical for learning good representations \cite{tian2020makes}, and unlike vision, the pool of augmentations that work well for graphs is limited. Half-Hop serves as a simple and useful addition to this existing toolkit of graph augmentations that can work on a wide range of datasets. Although our empirical investigations have provided evidence of the robustness of our approach, further studies are needed to understand the types of invariances introduced into the representation under Half-Hop and, to investigate how this augmentation performs for downstream tasks, such as link prediction or graph classification. 

\section*{Acknowledgments}
We would like to thank Mohammad Gheshlaghi Azar and Bernardo Avila Pires for their feedback on the work. This project was supported by NIH award 1R01EB029852-01, NSF awards IIS-2212182 and IIS-2146072, as well as generous gifts from the Alfred Sloan Foundation (ELD), the McKnight Foundation (MA, ELD), and the CIFAR Azrieli Global Scholars Program (ELD).

\bibliographystyle{icml2023}
\bibliography{main}

\newpage
\onecolumn
\appendix
\setcounter{figure}{0} 
\setcounter{table}{0} 

{\Large \bf Appendix} 

\section{Generalization Analysis}
\label{appendix:analysis}
\subsection{Problem Setup}

We adopt the model developed in \cite{keriven2022not} for our analysis. We invite the reader to refer to their work for more details. In this section, we add the necessary context (assumptions and notations) to set up our own result presented in the section A.2.

\textbf{Semi-supervised node regression.}~
We consider semi-supervised learning on an undirected graph of $n$ nodes. We observe the entire graph, encoded by the adjacency matrix $W$, and the node features $x_1,x_2,\dots,x_n$. Among the nodes, $n_{\text{tr}}$ nodes' labels are observed while $n_{\text{te}}=n-n_{\text{tr}}$ nodes are without labels and used for test. We stack the labels and features as rows of the matrices $Y$ and $X$, {\color{black} and denote the training labels/features as $Y_{\text{tr}}$ and $X_{\text{tr}}$ and the testing labels/labels as $Y_{\text{te}}$ and $X_{\text{te}}$.} 

{\color{black} Instead of predicting the labels using the node features alone, inference can be improved by leveraging the connectivity of the graph. We consider a linear message passing neural network (MPNN). $k$ rounds of message passing are applied according to the adjacency matrix $W$, which results in the {\em smoothing} of the node embeddings. The node labels are then predicted from the updated node embeddings.}

After multiple rounds of message passing, we then estimate the underlying model parameters from the updated node embeddings as:
\begin{align}\label{est}
 \hat{\beta}^{(k)} = \operatorname{argmin}_\beta \frac{1}{2 n_{\mathrm{tr}}}\left\|Y_{\mathrm{tr}}-X_{\mathrm{tr}}^{(k)} \beta\right\|^2+\gamma\|\beta\|^2,
\end{align}
{\color{black}where $X_{\mathrm{tr}}^{(k)}$ are the  embeddings of the training nodes obtained after k rounds of message passing, and $\gamma > 0$ is a ridge penalty.}

The test risk for our semi-supervised setting can be written as 
\begin{align}
 \mathcal{R}^{(k)} = n_{\mathrm{te}}^{-1}\left\|Y_{\mathrm{te}}-X_{\mathrm{te}}^{(k)} \hat{\beta}^{(k)}\right\|^2, 
\end{align}
 where $X_{\mathrm{te}}^{(k)}$ are the feature embeddings of the test nodes after k rounds of message passing, and $\hat{\beta}^{(k)}$ is the estimator in \eqref{est} learned from the training data.

Before we present our analysis, we introduce the following technical function to simplify our later expressions for the risk.

{\bf Definition 1. 
\cite{keriven2022not}}~For any symmetric positive semi-definite input matrix $S \in R^{d \times d}$, we define the function
\vspace{-3mm}
\begin{equation*}
R_{\text {reg. }}(S) \stackrel{\textrm { def. }}{=}\left(\Sigma^{\frac{1}{2}} \beta^{\star}\right)^{\top} K \left(\Sigma^{\frac{1}{2}} \beta^{*}\right) \in \mathbb{R}_{+}
\vspace{-4mm}
\end{equation*}
\textrm{where}~$K = \left(\mathrm{Id}-S^{\frac{1}{2}} M\left(\gamma \mathrm{Id}+M^{\top} S M\right)^{-1} M^{\top} S^{\frac{1}{2}}\right)^2$, and $M,~ \beta^*,~\Sigma$ are introduced in the model assumptions.

{\color{black}
{\bf Previous risk results on ordinary MPNN.}
In \cite{keriven2022not}, they show that the risk associated with the ridge regression operator without any message passing of raw features can be approximated by $$\mathcal{R}^{(0)} \simeq R_{\text {reg. }}\left( \Sigma \right),$$ whereas the risk after $k$ rounds of message passing is approximately,
$$\mathcal{R}^{(k)}\simeq R_{\text {reg. }}\left( A^{2k}\Sigma \right),$$ where $A = (\mathrm{Id} + \Sigma^{-1})^{-1}$.

One key implication of this result is that multiple rounds message passing induces a form of spectrum smoothing. This can in some cases, for a small number of steps, induce a helpful form of regularization. As evidenced in above equations, the MPNN, with $k$ rounds of message passing, effectively modifies the key dynamic component of the risk that depends on the covariance of raw data features, from $\Sigma$ to $A^{2k}\Sigma$.

In \cite{keriven2022not}, they show that analyzing the eigenvalues provides further insight into the smoothing phenomena. We note $\lambda_i$ an eigenvalue of the covariance matrix $\Sigma$, and $\lambda_i^{(k)}$ the eigenvalues after $k$ round of smoothing. \cite{keriven2022not}  shows that while large eigenvalues mostly maintain their magnitudes $\lambda_i^{(k)}\sim \lambda_i$, small eigenvalues decay exponentially $\lambda_i^{(k)}\sim \lambda_i^{2k+1}$. In other words, small eigenvalues decay faster than large eigenvalues. In the case where $\beta^*$ is aligned with eigenvectors with small eigenvalues, this can introduce harmful bias into the estimator. This framework enables us to understand how and when smoothing becomes harmful (over-smoothing). The rapid decay of small eigenvalues can be attributed to this phenomena. In the following, we show that Half-Hop slows down this decay and effectively enables the more graceful smoothing that we observe empirically in Figure 3.

\subsection{Main Result}
To compare the test risk for the original graph vs. when we apply Half-Hop, we use the directed variant of Half-Hop: $\mathrm{HH}^{(1)}$ described in Appendix \ref{variants}. In this variant, there is no backward edge from the target node to the slow node. 

\textbf{Theorem 1.}   The test risk after $k\in\{1,3,5,\dots\}$ rounds of message passing with \textit {Half-Hop} are: 
\begin{align*}
\mathcal{R}^{(k)}_{\text{HH}_{\alpha}}&\simeq R_{\text {reg. }}\left(\frac{1}{2}A^{k-1}\left(\textrm{I}+\left((1-\alpha) \textrm{I}+\alpha A\right)^2\right)\Sigma\right),
\end{align*}
where $A= (\mathrm{I} + \Sigma^{-1})^{-1}$.
}

Through Theorem 1, we can inspect how Half-Hop changes the rate of smoothing in the graph and can change the underlying spectral properties of our features. Similar to the original graph, large eigenvalues preserve most of their magnitude through message passing $\lambda^{(k)}\sim (1+((1-\alpha)+\alpha\lambda)^2)\lambda$; However, small eigenvalues decay as $\lambda^{(k)}\sim (1+(1-\alpha)^2)\lambda^{k+1}$. From these observations, we see that the decay rate of small eigenvalues is halved compared with ordinary MPNN without Half-Hop. 

{\color{black}
\textbf{Proof.} 
The basic idea behind our proof is similar to \cite{keriven2022not}, where we use matrix concentrations to approximate the node features after multiple rounds of message passing. 
Following the proof of Theorem 4 in \cite{keriven2022not}, the regression risk of $\mathcal{R}^{(k)}_{\text{HH}_{\alpha}}$ is approximated by $R_{\text {reg. }}(\Sigma')$ where $\Sigma'$ approximates  the covariance of the node features after $k$ rounds of message passing. Hence, the proof boils down to calculating $\Sigma'$.

With Half-Hop, we introduce new nodes that we call slow nodes. In the augmented graph, there are two types of nodes: 1. the original nodes whose features we denote as $x_i \in \mathbb{R}^d$ and 2. the slow nodes, whose features we denote as $\tilde{x}_k$. When Half-Hop is applied, the path from original node $v_j$ to original node $v_i$ is replace by a path from original node $v_j$ to slow node $\nu_k$ to original node $v_i$. We note $\mathcal{V}_i$ the set of $(j,k)$ paths that lead to $i$. We will use the superscript to denote the number of steps for message passing. 

Let's consider the first message passing round, the original node $i$ will received messages from the slow nodes connected to it:
\begin{align*}
    \forall i, x_i^{(1)} &= \sum_{(j, k)\in \mathcal{V}_i} a_{ij} \tilde{x}_k^{(0)} \\
    & = \sum_{(j, k)\in \mathcal{V}_i} a_{ij} \left ((1-\alpha) x_i^{(0)} + \alpha x_j^{(0)}\right )\\
    & = (1-\alpha) x_i^{(0)} \left (\sum_{(j, k)\in \mathcal{V}_i} a_{ij} \right ) + \alpha \cdot \sum_{(j, k)\in \mathcal{V}_i} a_{ij} x_j^{(0)}\\
    &= (1-\alpha) x_i^{(0)} + \alpha \cdot \sum_{(j, k)\in \mathcal{V}_i} a_{ij} x_j^{(0)}
\end{align*}

$x_i$ denotes an i.i.d. sample from the underlying latent generative model. Now if we use Lemma 1 in \cite{keriven2022not} and leverage our assumption on the Gaussianity of our features and a large number of nodes, it holds with high probability, that the features after one round of message passing can be approximated as:
\begin{align}\label{f1}
    \forall i, x_i^{(1)} \simeq  (1-\alpha)x_i + \alpha Ax_i
\end{align}
Note that in the original analysis, they prove that $x^{(1)} \simeq A x^{(0)}$. Our model, however, scales the messages from the rest of the network by $\alpha$ while the self-embedding is preserved and scaled by $1-\alpha$.

The slow nodes are also updated, since they have a single incoming edge, they simply copy the feature of their corresponding source node:
\begin{align}
    \forall i, \forall (j, k) \in \mathcal{V}_i, \ \ \ \tilde{x}_k^{(1)} = x_j^{(0)}
\end{align}

Let's perform a second round of message passing, we apply the same process again. The original nodes are updated as follows:
\begin{align*}
\forall i, x_i^{(2)} &= \sum_{(j, k)\in \mathcal{V}_i} a_{ij} \tilde{x}_k^{(1)} \\
&= \sum_{(j, k)\in \mathcal{V}_i} a_{ij} x_j^{(0)}\\
& \simeq A x_i^{(0)}
\end{align*}

We note that after the second round of message passing, the feature embeddings of the original nodes, are identical to the feature embeddings we would have obtained without HalfHop after a single step of message passing.

The slow nodes receive a copy of the embedding of their source node:
\begin{align*}
    \forall i, \forall (j, k) \in \mathcal{V}_i, \ \ \ \tilde{x}_k^{(2)} = x_j^{(1)} \simeq (1-\alpha)x_j + \alpha Ax_j
\end{align*}

We can write a general formula where at each step, the original nodes receive and aggregate messages from the slow nodes (we note this operation $\mathrm{AGG}$). The slow nodes are updated based on the source node they are connected to. We write this as:
\begin{align}\label{dyn}
    x_\text{i}^{(t+1)}=\textrm{AGG}(\tilde{x}_k^{(t)}),~~\tilde{x}_k^{(t+1)} = x_j^{(t)}.
\end{align}

We can now apply the recursive formulas \eqref{dyn} iteratively. This assumes that we can approximate the AGG operation with a multiplication of $A$ for Half-Hop. This is due to the underlying model assumptions which imply that the node feature distribution remains the same after a linear combination of i.i.d. Gaussian variables and the node features are approximately independent when $n$ is large.
Hence, if we continue applying the recursive formula and replace each AGG operation over $x$ by $Ax$, then for any $k\geq 0$, mathematical induction yield $$x_i^{(2t+1)}=\alpha A^{(t+1)} x_i + (1-\alpha)A^{(t)}x_i, \hspace{1mm} \tilde{x}_k^{(2t+1)} = A^tx_j.$$ Now we complete the proof by recalling that the covariance of the original node feature $x$ is $\Sigma$ and hence, the updated covariance after $k$ steps of message passing is:
\begin{equation}
\label{eq:HHcov}
\Sigma'=\frac{1}{2}A^{k-1}\left(\textrm{I}+\left((1-\alpha) \textrm{I}+\alpha A\right)^2\right)\Sigma.
\end{equation}

{$\square$ End of proof.}
}

{\color{black}
{\bf Remark.} 
The modified covariance term in Equation~\ref{eq:HHcov} consists of two main terms, the first being a component that smooths the original covariance with $A$ at a rate of $k-1$, which is roughly half the original rate of smoothing for the graph without Half-Hop. In addition to this first slower smoothing term, we also find a second contribution to the new covariance. The second modified smoothing term is $ ((1-\alpha) \textrm{Id}+\alpha A )^2 $, where in this case we see a uniform boosting of the covariance spectrum coming from the first term and weighted by (1-$\alpha$), and a second term coming from a rescaling of $A$. Thus, for small values of $\alpha$ we can interpret this as having a strong boosting of the  self-loops in the graph. We also confirm that this is indeed the case for our analysis of the receptive field for different values of $\alpha$ in Figure~\ref{fig:oversmoothing}.
}

\section{Ablations}
\label{appendix:ablations}

\subsection{Testing the directionality of edges added in Half-Hop}
\label{variants}

Recall that for an edge $e_{ij}$, Half-Hop introduces a new slow node $\nu_k$ along the edge from $v_i$ to $v_j$ as follows: $$\mathrm{HH}: \quad v_i \rightarrow \nu_k \leftrightarrow v_j.$$ 

In this experiment, we try alternative connectivity schemes. In the first variant, $\mathrm{HH}^{(1)}$, we do not introduce a backward edge that goes from the destination to the slow node:
$$\mathrm{HH}^{(1)}: \quad v_i \rightarrow \nu_k \rightarrow v_j.$$

The second variant, $\mathrm{HH}^{(2)}$, we add an edge going from the slow node to the source node, which has the effect of creating a path from $v_j$ to $v_i$ that might not exist in the original graph:
$$\mathrm{HH}^{(2)}: \quad v_i \leftrightarrow \nu_k \leftrightarrow v_j.$$

We assess the influence of the above connectivity schemes for the GCN model across three different datasets - Texas, Actor and Cornell, and we tabulate our results in Table \ref{table:ablation1}. We note significantly better performance for our proposed connectivity scheme $\mathrm{HH}$, that we adopt for Half-Hop. This happens to be the more intuitive solution since 1) it preserves the directionality of the original edge it splits and 2) it allows the slow node to communicate with both source and target nodes, as it act as the mediator in message passing.

\begin{table}[h!]
\centering
\caption{{\em Ablations of different connectivity motifs for slow nodes on heterophilic datasets.} We report the performance for the GCN model and the Half-Hop augmentation applied with each of the different connectivity schemes.\label{table:ablation1}}
\vspace{2mm}
\begin{tabular}{cccc}
\hline
Dataset    & $\mathrm{HH}$: $v_i \rightarrow \nu \leftrightarrow v_j$   & $\mathrm{HH}^{(1)}$: $v_i \rightarrow \nu \rightarrow v_j$& $\mathrm{HH}^{(2)} $: $v_i \leftrightarrow \nu \leftrightarrow v_j$ \\
\hline
Texas
& 71.71 $\pm$ 8.76
& 68.8  $\pm$ 6.50
& 58.47 $\pm$ 5.56\\
Actor        
& 33.35 $\pm$ 1.00
& 32.17 $\pm$ 0.84
& 31.93 $\pm$ 1.26\\
Cornell        
& 63.42 $\pm$ 5.62 
& 57.66 $\pm$ 6.89
& 42.16 $\pm$ 6.57\\
\hline
\end{tabular}
\end{table}

\subsection{Testing different initializations for the slow node}

When introducing slow nodes along edges, we use linear interpolation of the source and target nodes to initialize the features of the slow nodes.
In this experiment, we ablate the initialization used  for the slow node (Section \ref{sec:virtualnode}). We test two simpler alternatives:  1) `zero': All of the features are set to zero, 2) `random': We use a uniform distribution in the range of [0,1) to initialize the features of the slow node.
The results are presented in Table \ref{table:ablation2}, where we find that the `zero' and `random' initializations are too simple and hurt the performance of the model. Linear interpolation, on the other hand, comes as a natural scheme that mixes features from the real node distribution.

\begin{table}[h!]
\centering
\caption{{\em Ablations of different initialization schemes for slow nodes on heterophilic datasets} We report the performance for the GCN model and the Half-Hop augmentation applied with each of the different initialization schemes. \label{table:ablation2}}
\vspace{2mm}
\begin{tabular}{cccc}
\hline
Dataset &  linear interpolation & zero & random \\
\hline
Texas
& 72.88 $\pm$ 7.17
& 61.80 $\pm$ 5.91
& 53.33 $\pm$ 5.27\\
Actor
& 33.39 $\pm$ 1.29
& 28.93 $\pm$ 2.83
& 24.70 $\pm$ 1.16\\              
Cornell
& 63.33 $\pm$ 5.70
& 49.55 $\pm$ 7.06 
& 37.48 $\pm$ 6.93\\
\hline
\end{tabular}
\end{table}

\section{Details for Supervised Experiments}
\label{appendix:supex}

\textbf{Homophily.} In this work, we follow the definition of \emph{node homophily ratio} as used in \cite{Pei2020Geom-GCN} given by the formula:
\begin{equation*}
\frac{1}{|\mathcal{V}|}\sum_{v \in \mathcal{V}} \frac{|\{(v, w) : w \in \mathcal{N}(v) \wedge y_v = y_w\}|}{|\mathcal{N}(v)|},
\end{equation*}
where $\mathcal{V}$ denotes the set of all nodes in the graph, $\mathcal{N}(v)$ denotes all the neighbors of an arbitrary node $v$, and $y_v$ denotes the class membership of the node $v \in \mathcal{V}$.

We classify datasets into \emph{homophilic datasets} and \emph{heterophilic datasets} based on the homophily score: datasets with homophily $\ge$ 0.5 are classified as \emph{homophilic datasets} and \emph{heterophilic datasets} otherwise.

\subsection{Homophilic Datasets}

We use five real-world datasets, Amazon Computers and Amazon Photos \cite{amazon}, Coauthor CS and Coauthor Physics \cite{coauthor} and WikiCS \cite{mernyei2020wiki}. Key statistics for the different datasets are listed in Table \ref{table:homo_datasets}.

\begin{table*}[!h]
\centering
\caption{{\em  Statistics of homophilic datasets used in our experiments.}}
\label{table:homo_datasets}
\vspace{3mm}
\resizebox{0.75\textwidth}{!}{
\begin{tabular}{l|cccccc}
\hline
 & Nodes & Edges& Features & Classes & Node Homophily Ratio \\
 \hline
 Amazon Photos  & 7,650 & 119,081  & 745 & 8 & 0.8365  \\
 Amazon Computers & 13,752 & 245,861 & 767 & 10 & 0.7853 \\
 Coauthor CS & 18,333 & 81,894  & 6,805 & 15 & 0.8320 \\
 Coauthor Physics & 34,493 & 247,962  & 8,415 & 5 & 0.9153 \\
 Wiki CS & 11,701 & 216,123  & 300 & 10 & 0.6588 \\
\hline
\end{tabular}}
\end{table*}

The experimental setup follows that of \cite{math10081262}, where we split the dataset into development and test sets. All the hyperparameter tuning is done on the development set and the best models are evaluated on the test set. The runs are averaged over 20 random splits to minimize noise.
We follow a 60:20:20\% train/val/test split for the Amazon and Coauthor datasets, and 20 pre-split masks provided in the WikiCS dataset.

\subsection{Heterophilic Datasets} \label{appendix:hetero}

We use five real-world datasets with graphs that have a homophily level $\le$ 0.30, Texas, Wisconsin, Actor, Chameleon and Cornell \cite{heterophily}.
Key statistics for the different datasets are listed in Table \ref{table:hetero_datasets}.

\begin{table*}[!h]
\centering
\caption{{\em  Statistics of heterophillic datasets used in our experiments.}}
\label{table:hetero_datasets}
\vspace{3mm}
\resizebox{0.55\textwidth}{!}{
\begin{tabular}{l|cccc}
\hline
 & Nodes & Edges & Classes & Node Homophily Ratio\\
 \hline
Texas & 183 & 295 & 5 & 0.11 \\
Wisconsin & 251 & 488 & 5 & 0.21\\
Film & 7,600 & 26,752 & 5 & 0.22\\
Squirrel & 5,201 & 198,493 & 5 & 0.22 \\
Chameleon & 2,277 & 31,421 & 5 & 0.23\\
Cornell & 183 & 280 & 5 & 0.30\\
\hline
\end{tabular}}
\end{table*}

We follow the experimental setup in \cite{Pei2020Geom-GCN}, we use the same 10 train/val/test splits that are provided. We also perform similar hyperparameter tuning using randomized grid search using only the train and validation sets. Once we find the best model, we report the accuracy on the test set, which is only seen once. We include the final hyperparameters of the best models for each architecture and dataset in Table \ref{tab:hp_hetero}.

\begin{table*}[th]
\centering
\caption{{\em Best hyperparameters found using the validation set in our experiments on heterophilic datasets.} \label{tab:hp_hetero}}
\vspace{3mm}
\resizebox{0.75\textwidth}{!}{
\begin{tabular}{l|c|ccccc|cc}
\hline
Dataset & Model & lr & weight decay & depth & hidden & dropout & $\alpha$ & p   \\
\hline
\multirow{3}{*}{Texas} & HH-GCN & 0.0291 & 0.0096 & 2 & 64 & 0.8058 & 0.0043 & 0.9526\\ 
& HH-GraphSAGE &  0.0170 & 0.0053 & 2 & 64 & 0.1967 & 0.9397 & 0.7140 \\
& HH-GAT & 0.0328 & 0.0066 & 2 & 32 & 0.1288  & 0.0902 & 0.9841 \\
\hline
\multirow{3}{*}{Wisconsin} & HH-GCN & 0.0105 & 0.0002 & 3 & 128 & 0.6612 & 0.9937 & 0.7140 \\
 & HH-GraphSAGE &  0.0202 & 0.0042 & 3 & 64 & 0.3462 & 0.0100 & 0.6177 \\  
& HH-GAT &  0.0539 & 0.0068 & 3 & 16 & 0.2141 & 0.0026 & 0.9797 \\
\hline
\multirow{3}{*}{Actor} & HH-GCN & 0.0313 & 0.0087 & 3 & 64 & 0.5511 & 0.0369 & 0.5466 \\
& HH-GraphSAGE &  0.0133 & 0.0090 & 3 & 32 & 0.3737 & 0.0116 & 0.8368 \\ 
& HH-GAT &  0.0009 & 0.0001 & 3 & 128 & 0.8708 & 0.0549 & 0.9594 \\
\hline
\multirow{3}{*}{Squirrel} & HH-GCN &  0.0053 & 0.0001 & 3 & 128 & 0.2455 & 0.0145 & 0.8257 \\
 & HH-GraphSAGE & 0.0296 & 0.0001 & 2 & 128 & 0.8668 & 0.9474 & 0.5198\\
 & HH-GAT &  0.0027 & 0.0001 & 3 & 64 & 0.5131 & 0.9277 & 0.1549 \\
\hline
\multirow{3}{*}{Chameleon}  & HH-GCN &  0.0318 & 0.0057 & 2 & 128 & 0.8040 & 0.0510 & 0.9986 \\
& HH-GraphSAGE & 0.0225 & 0.0001 & 2 & 32 & 0.7175 & 0.9834 & 0.6226\\
& HH-GAT &  0.0012 & 0.0008 & 3 & 64 & 0.0439 & 0.9766 & 0.9386\\
\hline
\multirow{3}{*}{Cornell}   & HH-GCN &  0.0505 & 0.0055 & 2 & 32 & 0.4123 & 0.0145 & 0.9660 \\
& HH-GraphSAGE &  0.0697 & 0.0018 & 2 & 64 & 0.0697 & 0.8807 & 0.5660 \\
 & HH-GAT &  0.0572 & 0.0070 & 2 & 64 & 0.0572 & 0.0710 & 0.9979 \\
\hline
\end{tabular}}
\end{table*}

\section{Details for Self-supervised Experiments}

We use the same real-world datasets (Amazon-Photos, Amazon-Computers, Coauthor-CS and Coauthor-Physics) used in the supervised setting. The full graph is used during pre-training (transductive task), then during linear evaluation, the graph is split into train/val/test with 10:10:80\% of the nodes respectively. This setup follows \cite{thakoor2022large}. The method of evaluation follows the linear evaluation protocol \cite{dgi}, where the weights of the model are frozen and a linear classifier is trained on top of the learned representations (without propagating gradients to the encoder). We use an $l_2$-regularized Logistic Regression with a \emph{liblinear} solver from the Scikit-learn library \cite{sklearn}. 

For all of the experiments, we use the same hyperparameters as GRACE \cite{grace} and BGRL \cite{thakoor2022large} notably, the edge-dropout and feature-dropout hyperparameters defined for each view: we have the \emph{edge-masking probabilities} ($p_{e, 1}$, $p_{e, 2}$), and the \emph{feature-masking probabilities}  ($p_{f, 1}$, $p_{f, 2}$). For Half-Hop the hyperparameters we introduce are the \emph{Half-Hop probabilities} ($p_{hh, 1}$, $p_{hh, 2}$) and the linear interpolation coefficients used to initialize the slow nodes ($\alpha_1$, $\alpha_2$). We report all these numbers in Table \ref{tab:bgrl_hp}.

\begin{table}
\centering
\caption{Augmentation hyperparameters used to train HH-BGRL, HH-GRACE. \label{tab:bgrl_hp}}
\vspace{3mm}
\resizebox{0.65\textwidth}{!}{
\begin{tabular}{c|ccccc}
\hline
Aug. Hyperparameters & Am. Comp. & Am. Photos & Co. CS & Co. Phy & WikiCS \\
\hline
$p_{hh, 1}$  &    0.75   &    0.75    &   0.75  &  0.75   &  0.75  \\
$p_{hh, 2}$  &    0.75   &    0.75    &   0.75  &  0.75   &  0.75  \\
$\alpha_{1}$   &     0.50   &     0.50    &   0.50  &   0.50   &   0.50  \\
$\alpha_{2}$   &     0.50   &     0.50    &   0.50  &   0.50   &   0.50  \\
\hline
$p_{f, 1}$      &     0.20   &    0.10     &   0.30  &   0.10   &   0.20  \\
$p_{f, 2}$      &     0.10   &    0.20     &   0.40  &   0.40   &   0.10  \\
$p_{e, 1}$         &     0.50   &    0.40     &   0.30  &   0.40   &   0.20  \\
$p_{e, 2}$          &     0.40   &    0.10     &   0.20  &   0.10   &   0.30  \\
\hline
\end{tabular}
}
\end{table}

 \label{appendix:hp}

\section{Comparisons with other GNNs}

There have been multiple modifications on top of traditional GNN architectures to optimize for the task of heterophilic node classification. 
In Table \ref{fig:GNNcomparisons} we detail the different architectural components, losses, and design choices that are used to improve performance in heterophilic datasets. In the table, we breakdown the different components of popular methods that we compare with in the main text, including: MixHop \cite{abu2019mixhop},  (ii) GGCN \cite{yan2021two}, and (iii) $H_2$GCN \cite{zhu2020beyond}. 

\renewcommand{\arraystretch}{1.2}
\begin{table*}[h!]
\centering
\begin{tabular}{|c|c|c|c|c|}  
    \cline{2-5}
    \multicolumn{1}{c|}{} & {\small Higher-order neighbors} & {\small Weights for self-loops} & {\small Concat across layers}& {\small Dynamic gating} \\
    \hline
      GCN & \xmark & \xmark & \xmark & \xmark \\
      \hline
     SAGE & \xmark & \cmark & \xmark & \xmark  \\
    \hline
    MixHop & \cmark & \xmark & \cmark & \xmark  \\
    \hline  
    GGCN & \xmark & \xmark & \xmark & \cmark  \\
    \hline
    H$_2$GCN & \cmark & \cmark & \cmark & \xmark \\
    \hline
     HH-GCN & \xmark & \xmark & \xmark & \xmark \\
     \hline
     HH-SAGE & \xmark & \cmark & \xmark & \xmark \\
     \hline
\end{tabular}

\caption{Different components used in graph neural networks optimized for heterophilic node classification. From left to right, we show methods that incorporate additional information from higher-order neighbors, separate weights for self-loops, and other additional components. Here, we observe the fact that Half-Hop is lightweight and doesn't require extra components in the loss and also doesn't explicitly compute separate weights for self-loops. \label{fig:GNNcomparisons}}
\end{table*}


\section{Visualization of the embedding space across layers}
\label{app:latentviz}

\begin{figure}[!h]
\centering
\includegraphics[width=\textwidth, trim={0 0.8cm 0 0.7cm},clip]{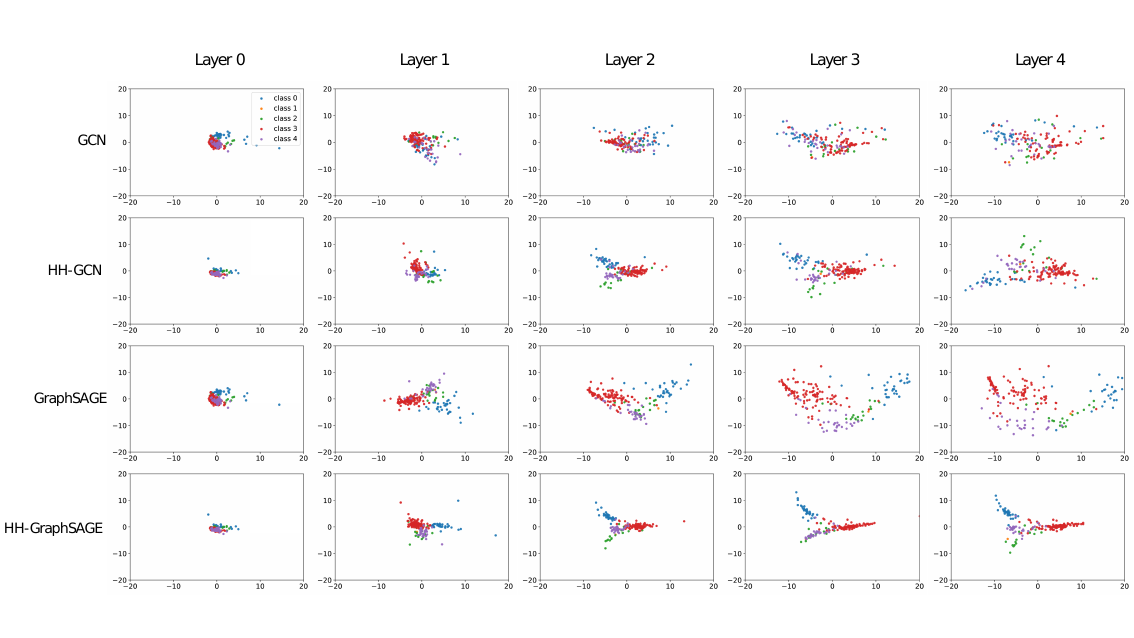}
\vspace{-12pt}
\caption{Latent space visualizations with and without Half-Hop for Texas dataset. We can note how similar data points cluster closer together in the cases where Half-Hop is applied. As class 1 is very rare, it is not clearly visible in the visualization. 'Layer 0' denotes the latent space before the GNN layers are applied.\label{table:latentvis_texas}}
\end{figure}

\begin{figure}
\centering
\includegraphics[width=\textwidth, trim={0 0.8cm 0 0.7cm},clip]{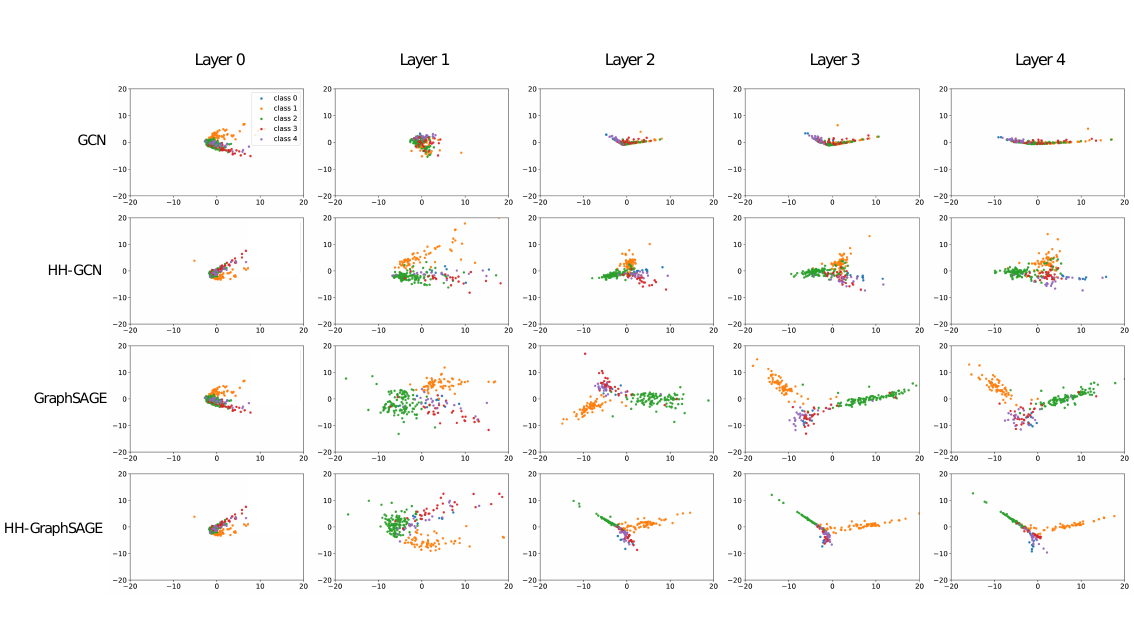}
\vspace{-12pt}
\caption{Latent space visualizations with and without Half-Hop for Wisconsin dataset. Though the color scheme is the same, it appears different from that of Figure \ref{table:latentvis_texas} due to the difference in the distribution of nodes across classes in Texas and Wisconsin datasets.\label{table:latentvis_wisconsin}}
\end{figure}

\begin{figure}
\centering
\includegraphics[width=\textwidth, trim={0 0.8cm 0 0.7cm},clip]{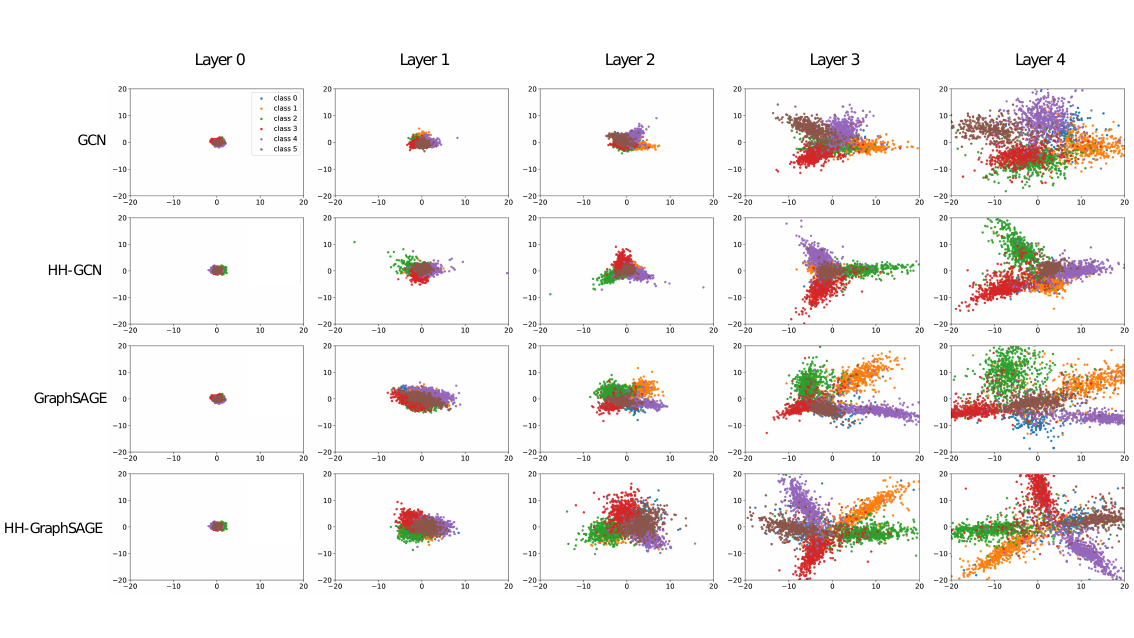}
\vspace{-12pt}
\caption{Latent space visualizations with and without Half-Hop for Citeseer dataset. We can again note how in the cases where Half-Hop is applied, similar latents seem to cluster closer together. Citeseer has a larger number of nodes compared to Texas and Wisconsin datasets and hence the latents in these visualizations appear to be packed more densely in comparison.\label{table:latentvis_citeseer}}
\end{figure}

In this experiment, we visualize the latent space of node embeddings across various layers, with and without Half-Hop (Figures~\ref{table:latentvis_texas}, \ref{table:latentvis_wisconsin} and \ref{table:latentvis_citeseer}) using GCN and GraphSAGE encoders.

Observing the latent space visualizations, we can make a few interesting observations. Upon applying Half-Hop, the embeddings of the same class appear to aggregate together better while the embeddings of different classes seem to maximally distance themselves from each other and from the center of the latent space. Though this is slightly observed in all cases, this can be clearly noted in the case of GraphSAGE + Half-Hop for the Citeseer dataset in Figure \ref{table:latentvis_citeseer}. Visualizations of latents for vanilla GCN (without Half-Hop) indicate poor class separation (Figure \ref{table:latentvis_texas} and Figure \ref{table:latentvis_wisconsin}) and this can be seen reflected directly in vanilla GCN's performance as noted in Table \ref{tab:hetero} in the paper. On a more general note, class separation appears best after the second and third layers as observed across Figures \ref{table:latentvis_texas}, \ref{table:latentvis_wisconsin} and \ref{table:latentvis_citeseer}. This is similar to what is observed in terms of raw performance - the best-performing models are 2 or 3 layers deep.

\end{document}